# Dynamic Trees: A Structured Variational Method Giving Efficient Propagation Rules

**Amos J Storkey**
Institute of Adaptive and Neural Computation
Division of Informatics, University of Edinburgh
5 Forrest Hill, Edinburgh UK
a.storkey@ed.ac.uk

## Abstract

Dynamic trees are mixtures of tree structured belief networks. They solve some of the problems of fixed tree networks at the cost of making exact inference intractable. For this reason approximate methods such as sampling or mean field approaches have been used. However, mean field approximations assume a factorised distribution over node states. Such a distribution seems unlikely in the posterior, as nodes are highly correlated in the prior. Here a structured variational approach is used, where the posterior distribution over the non-evidential nodes is itself approximated by a dynamic tree. It turns out that this form can be used tractably and efficiently. The result is a set of update rules which can propagate information through the network to obtain both a full variational approximation, and the relevant marginals.

The propagation rules are more efficient than the mean field approach and give noticeable quantitative and qualitative improvement in the inference. The marginals calculated give better approximations to the posterior than loopy propagation on a small toy problem.

## 1 Introduction

The main subject matter of this paper is the dynamic tree model, which was introduced in [12] as a model of images. The dynamic tree is a particularly versatile hierarchical generative model. It offers improvements on fixed tree belief networks such as quadtree networks. Quadtree networks have proven useful for image segmentation, and allow exact probabilistic inference using belief propagation [13, 2]. However quadtrees suffer from model non-homogeneity: quadtrees produce blocky artefacts due to the fact that two (spatially) adjacent leaf nodes might only be path-connected through a vertex far up the tree hierarchy.

Dynamic trees reduce this problem by removing the constraints of the fixed structure. They allow each node of the network to 'choose' its parent. The result is a mixture of tree networks. Because of this mixture, exact inference is no longer feasible; its computational cost is exponential in the number of nodes.

Dynamic tree-like structures have also been used for hierarchical clustering [11], and can be considered for use in any situation where tree structured belief networks are used, but are considered to be too rigid.

## 2 Dynamic Trees

### 2.1 Theory

There are two components to a dynamic tree model: a prior distribution of possible tree architectures, and the conditional probabilities of each node given its parents and the tree architecture. For the standard dynamic tree model the nodes are arranged in layers, and each node 'chooses' its parent from those in the layer above. Nodes are given positional values, and prospective parents closer to a given node are given higher probabilities of being chosen by that node. The nodes themselves are given discrete states, which are used to denote which particular object is being represented, and to ultimately determine what pixel configuration is seen at the leaf nodes of the network.

Consider a set $V$ of nodes $i = 1, 2, \ldots, n$ and a set $S$ of possible states $1, 2 \ldots, m$ of each node. Let $Z = \{z_{ij}\}$ denote the set of possible directed tree structures over these nodes, where $z_{ij}$ is an indicator. $z_{ij} = 1$ denotes the fact that node $j$ is the parent of node $i$. Finally let $X = \{x_i^k\}$ represent the state of the nodes: $x_i^k = 1$ if node $i$ is in state $k$, and is zero otherwise.

A dynamic tree can then be represented by a prior over the possible trees $P(Z)$, and a prior over the network states given a particular tree structure $P(X|Z)$. We assume that the prior over $Z$ factorises: each node 'chooses' a parent from a set of possible parents, and chooses independently of other nodes. Hence $P(Z) = \prod_{ij} \rho_{ij}^{z_{ij}}$, where $\rho_{ij}$ is the probability that node $i$ chooses parent $j$. In defining $P(Z)$ here, we order the nodes into layers; nodes can only choose a parent from the layer above.

The prior over the network states is given by the conditional probability tables of the network. Whatever the value of $Z$, the conditional probability of one node given another is always the same if they are connected. The conditional probability $P_{ij}^{kl}$ defines the probability of moving from a state $l$ to state $k$ when traversing a network link from $j$ to $i$.

With these prior forms, the joint prior distribution can be written as

$$P(Z, X) = \prod_{i,j=1}^{n} \rho_{ij}^{z_{ij}} \prod_{kl} [P_{ij}^{kl}]^{x_i^k x_j^l z_{ij}} \quad (1)$$

where the indicator variables are simply used to pick out the correct probabilities.

The nodes (vertices) are split into a set $V^E$ and a set $V^H$ of evidential and non-evidential (hidden) nodes respectively. Likewise the corresponding node state indicator variables are denoted by $X^E$ and $X^H$ respectively. The posterior distribution of the dynamic tree can then be written as $P(Z, X^H|X^E) = P(Z, X)/P(X^E)$. Usually the evidential nodes are the leaf nodes of the network.

Given some data, which we use to instantiate the leaf (evidential) nodes of the network, we want information about the posterior distribution of the tree structures and the nodes of the network. Calculating these posterior probabilities exactly is infeasible because it would involve a belief propagation for each tree in the mixture, and the number of trees scales exponentially in the square of the number of nodes. Therefore we will need to resort to approximate methods, such as those outlined in section 3

Note that this prior is a mixture of trees. The $Z$ variables can be integrated out to get another belief network representation of this prior, one where each node is the child of all the nodes in the layer above. Sometimes it will be useful to think in these terms, and henceforth this form will be referred to as the multi-parent belief network representation.

## 2.2 Intuition behind dynamic trees

Dynamic trees are predominantly used as models for images. They take their impetus from a generative approach. Instead of primarily looking for features or characteristics within an image, and then trying to use this information for segmentation, recognition or some other purpose, a generative approach starts with asking what we know about how the image came about. This gives us some prior knowledge of what we might expect to see in an image. We use this prior knowledge to build what is called a 'generative model'. The perceived image is then used to refine that prior knowledge to provide a reasonable model (or set of models) for *that* particular image.

Images can be seen as being constructed from many different objects. These objects often have component parts, and these parts contain other substructures and so on. The image is a two dimensional pixellation of the scene described by these various structural elements.

Given this, it would appear sensible to model objects hierarchically. A simple deterministic model will not capture the variability in object structure between different images or parts of images, thus a probabilistic model is more appropriate.

Using a tree-structured directed graph (see figure 1) is a good way to define a hierarchical probabilistic model. In such a structure, nodes are used to represent the different variables, and the probability of a node being in some state is given in terms of the possible states of its parent (each directed edge of the graph goes from a *parent* node to a *child* node). In other words whether the parent is in a certain state or not affects the state probability of the child node. This is reasonable as any dependence between objects would only occur through the component-subcomponent relationship. Hence distinct unrelated objects would be probabilistically independent in the model.

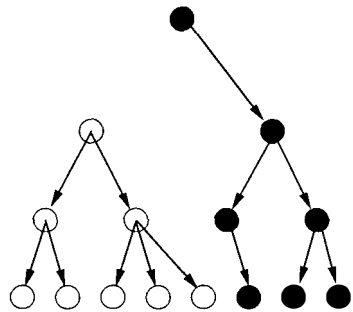

Figure 1: A disjointed tree structured directed graph

The problem with using a fixed tree structure is that the inherent organisation of different object scenes would not be correctly represented by the same hierarchical relationships. Hence the prior image model must incorporate some method of combining a host of sensible tree structures. Dynamic trees do just that. They adopt a method of forming a mixture of trees which is both flexible, but also encodes the requirement that the component parts of an object must be in some way spatially local to that object.

This last characteristic is the primary feature of dynamic trees. And while this component is vital for such an approach to modelling images, its introduction prevents the use of belief propagation for tractable exact probabilistic inference. That intractability is the problem with which this paper is involved.

### 2.3 Inference

How should inference in dynamic trees be tackled? Using simulated annealing to find the maximum a posteriori tree $Z^* = argmaxP(Z|X^E)$ was discussed in [12]. This was compared with the mean field approach in [1]. The mean field approach seemed to give better approximations to the posterior, while being significantly faster. However the mean field approach suffers from an unlikely choice of approximating distribution; it requires that the posterior be well represented by a factorised distribution over the nodes. Because the tree structure imposes strong conditional dependencies between connected nodes, we would also expect these dependencies to occur in the posterior. Though the mean field is able to capture some correlation between nodes by biasing the mean values towards the same class, it cannot capture these conditional dependencies.

Variational approximation methods have been discussed in [7, 4] and elsewhere. The mean field is the simplest form of variational approximation, but if it is tractable to use a more general form of distribution then it might be possible to capture more of the structure of the posterior distribution.

Loopy propagation is a different way of performing inference in belief networks. Its origin is in the belief propagation approach which Pearl advocated for singly connected networks [6]. Belief propagation exactly calculates marginal probabilities in singly connected networks. Theoretically, though, it is inappropriate to use it in multiply connected graphs. However Pearl noted that reasonable results could be achieved by doing just that. Since then various authors [9, 8, 3] have investigated the properties of loopy propagation and elaborated on its workings and accuracy. Loopy propagation could well be applied to networks such as the dynamic tree.

In this paper a more general variational form than the mean field approximation is proposed. Instead of approximating the posterior with a factorised distribution over the nodes, the whole posterior is represented by a dynamic tree. It turns out that this approach can be followed tractably and efficiently. It also captures many of the conditional dependencies in the posterior, and provides a closer approximation to the true posterior than both mean field and loopy propagation.

Whatever approximation method we choose, there are three important requirements. First we need some way of getting a measure of the likelihood of the model. This is vital for learning. Second we require some indication of the posterior *distribution* over tree structures. This will enable the image structure to be captured. Lastly we need some idea of the probabilities of finding different object labels at different parts of the structure.

### 2.4 Learning in dynamic trees

Because the variational method provides a lower bound to the log probability, we can use an expectation-maximization (EM) approach for learning the parameters of $P$ such as that described in [5]. The parameters of $P$ are chosen to minimize the sum of the variational free energies over all the elements of a dataset. This (local) minimization is straightforward because all the relevant derivatives can be calculated.

## 3 Variational approaches

Here we will quickly outline the general variational approach, and more specifically, the mean field method. The mean field approach for dynamic trees has been described in more detail in [1].

### 3.1 General outline

Explicit calculation of the posterior distribution is not viable. Therefore it seems sensible to find some reasonable approximation to the posterior instead. Suppose we denote this approximate distribution by $Q(Z, X^H)$. The variational approach says that the parameters of the distribution $Q(Z, X^H)$ should be adjusted to minimize the Kullback-Liebler (KL) divergence between the $Q$ distribution and the true posterior. If the $Q$ distribution becomes identical to the true posterior then the KL divergence is zero. Otherwise the KL divergence is greater than zero. The KL divergence is given by

$$KL(Q||P)$$
$$= \langle \log Q(Z, X^H) - \log P(Z, X^H|X^E) \rangle_{Q(Z,X^H)}$$
$$= \langle \log Q(Z, X^H) - \log P(Z, X) \rangle_Q + \log P(X^E) \quad (2)$$

where $\langle . \rangle$ denotes an expectation. Because $P(X^E)$ is constant with respect to changes in $Q$, minimizing the KL divergence corresponds to minimizing the first term on the right hand side of (2), called the variational free energy. Furthermore, because the KL divergence is always positive, this variational free energy gives an upper bound to $-\log P(X^E)$. This is a major benefit of the variational framework as it means that any parameters of the $P$ distribution can be updated using a form of EM algorithm [5].

A general approach to variational methods was discussed in [10]. However this approach produced variational approximations that scaled exponentially with the size of the largest clique. Unfortunately for this (and many other) models, the clique size can be very large. Here, for a network of $n$ nodes, the largest clique can contain about $n/4$ nodes (if each child chooses from all nodes in the layer above). Hence the largest clique will have at least $2^{n/4}$ states. Here we need some other approach which will still give tractable inference while keeping approximating models with large cliques.

The mean field is the simplest variational method. For dynamic trees, the mean field approach involves approximating the posterior with a distribution of the form $Q(Z)Q(X^H)$, where $Q(Z) = \prod_{ij} \mu_{ij}^{z_{ij}}$ factorises, and $Q(X^H)$ takes the factorising form $Q(X^H) = \prod (m_i^k)^{x_i^k}$. Here $\mu_{ij}$ and $m_i^k$ are parameters which need to be adapted to best fit the approximating distribution to the true posterior.

This mean field approximation causes problems. In [1] it was shown that spontaneous symmetry breaking polarises the distribution of the nodes far from the data. This occurs because many large conditional posterior dependencies are not captured by the form of $Q(X^H)$, and the resulting error has a significant effect on the approximate posterior distribution over tree structures.

### 3.2 A structured variational approach

It would seem appropriate to try to use some more general approximate form for $Q(X^H)$ in order to capture the conditional dependencies. Unfortunately with any significant structure in $Q(X^H)$ the variational calculation is not easily expressed or computed: for each node, it would involve expression and computation of the pairwise joint marginals of $Q(X^H)$ for all node pairs connected in some $Z$. If we want to make any improvement, we will have to go back to the beginning, and remove the assumption that $Q$ factorises into $Q(Z)Q(X^H)$.

A new approach is needed. Instead of assuming that $Q$ factorises into two parts, we use another dynamic tree (DT) as an approximating distribution. Though this might seem like a rather complicated approximation, it is in fact tractable.

In detail, the approximating $Q$ has the form $Q(Z, X^H) = Q(Z)Q(X^H|Z)$ where $Q(Z) = \prod_{ij} \mu_{ij}^{z_{ij}}$ is the same form of factorising distribution (with parameter $\mu$, to be optimised), but now the distribution over $X$ is dependent on the choice of tree and has the form

$$Q(X^H|Z) = \prod_{ij \in V^H} \prod_{kl} [Q_{ij}^{kl}]^{x_i^k x_j^l z_{ij}} \quad (3)$$

where the parameters $Q_{ij}^{kl}$ are conditional probabilities, and need to optimised.

This form of distribution, though significantly more general than the mean field, is still less general than the posterior, which would not have a factorised $P(Z|X^E)$, for example. However it is both possible to use it as a variational approximation, and possible to calculate its marginal distributions. So it satisfies our requirements for a more general but usable distribution.

### 3.3 Determining the parameters

To choose good parameters for the approximating distribution, the Kullback-Liebler (KL) divergence between the $Q(Z, X^H)$ distribution and the true posterior should be minimised. The KL divergence is of the form $KL(Q||P) = \log P(X^E) + V_Z(Q(Z)) + V(Q, P)$ where $V_Z(Q(Z)) = \sum_Z Q(Z)[\log Q(Z) - \log P(Z)]$ and

$$V(Q, P) = \sum_{Z, X^H} Q(Z) Q(X^H|Z) \log \frac{Q(X^H|Z)}{P(X|Z)}$$

When written out explicitly, $V_Z(Q(Z)) = \sum_{ij} \mu_{ij} \log(\mu_{ij}/\rho_{ij})$. Now let $m_i^k$ denote the mean values (under $Q(Z, X)$) of the indicator $x_i^k$ for non-evidential nodes.

For notational convenience we let $m_i^k$ equal the instantiated $x_i^k$ for the evidential nodes $i$, and propose additional degenerate connections for the same evidential nodes. The conditional probability matrix $Q_{ij}$ for such a connection from evidential node $i$ to some node $j$ will take the form $\mathbf{x}_i \mathbf{1}_m^T$. In other words the $ab$ element of matrix $Q_{ij}$ takes value $x_i^a$ whatever the values of $b$ and $j$. This simply says that these connections will always generate the data regardless of the state of the nodes in the layer above. These connections are fixed; they are not optimised over. When these additional nodes are incorporated into the variational free energy calculation, it can be seen that they make no contribution.

Having introduced these connections, we find that $V(Q, P)$ can be written in terms of its parameters more simply as a sum over all the nodes:

$$V(\{Q\}, \{\mu\}) = \sum_{ij \in V} \mu_{ij} \left[ \sum_{kl} Q_{ij}^{kl} m_j^l \log \frac{Q_{ij}^{kl}}{P_{ij}^{kl}} \right]$$

where the $m$'s are fully determined by the $Q$'s.

In the next few paragraphs the methods for optimising the $Q$'s and the $\mu$'s are outlined. These methods depend on the fact that any change in a connection parameter only impacts on the means of the state variables for nodes below that connection. This enables us to decompose the KL divergence into simple parts.

**Optimising Q** To obtain the best parameters we look at the derivatives of the KL divergence with respect to each of them. Considering the $Q$ parameters first of all,

$$\frac{\partial KL}{\partial Q^{ab}_{st}} = \frac{\partial V_{st}}{\partial Q^{ab}_{st}} + \sum_k \frac{\partial V^k_{d(s)}}{\partial m^k_s} \frac{\partial m^k_s}{\partial Q^{ab}_{st}} \quad (4)$$

where

$$V_{ij}(Q_{ij}, \mu_{ij}, \mathbf{m}_j) = \sum_{kl} \mu_{ij} \left[ Q^{kl}_{ij} m^l_j \log \frac{Q^{kl}_{ij}}{P^{kl}_{ij}} \right]$$

and $V^k_{d(s)}(\mu_{ij}, m^k_s, \{Q(i,j) | i < r(s)\}) =$

$$\sum_{i \in d(s), j} \sum_{kl} \mu_{ij} [Q^{kl}_{ij} m^l_j (\log Q^{kl}_{ij} - \log P^{kl}_{ij})]$$

Here $d(s)$ denotes the set of nodes in layers below $s$ (i.e. the descendants of $s$ in the multi-parent belief network representation). The $m^l_j$ terms are entirely dependent on the $m^k_s$ and the $Q$'s.

Most of the derivatives in (4) are straightforward to compute. The exception is $\frac{\partial V^k_{d(s)}}{\partial m^k_s}$. This can be obtained by propagating the derivatives from the layer below.

$$\frac{\partial V^k_{d(s)}}{\partial m^k_s} = \sum_{c \in c(s)} \left( \frac{\partial V_{cs}}{\partial m^k_s} + \sum_p \frac{\partial V^p_{d(c)}}{\partial m^p_c} \frac{\partial m^p_c}{\partial m^k_s} \right) \quad (5)$$

where $c(s)$ denotes the set of nodes in the layer immediately below $s$ (i.e. children of $s$ in the multi-parent belief network representation) The $Q$'s are updated from the bottom layer to the top layer by first propagating the derivatives (5):

$$T^k_s = \frac{\partial V_{d(s)}}{\partial m^k_s} = \sum_{c \in c(s)} \sum_g \mu_{cs} Q^{gk}_{cs} (\log \frac{Q^{gk}_{cs}}{P^{gk}_{cs}} + T^g_c) \quad (6)$$

and secondly optimising $\frac{\partial KL(Q||P)}{\partial Q^{ab}_{st}}$ with Lagrange multipliers to encode the probabilistic constraints on the $Q$'s. This gives

$$Q^{ab}_{st} = \frac{P^{ab}_{st} \exp(-T^a_s)}{\sum_a P^{ab}_{st} \exp(-T^a_s)} \quad (7)$$

for non leaf nodes $s$. Defining $\lambda^k_s$ to be $\exp(-T^k_s)$, and substituting the above form for $Q$ in to (6), we obtain

$$\lambda^k_s = \prod_{c \in c(s)} \left[ \sum_g P^{gk}_{cs} \lambda^g_c \right]^{\mu_{cs}} \quad (8)$$

In fact we get the same form for the $\lambda$ values propagated up from the leaf nodes if we initialise each leaf node $i$ with $\lambda^k_i = m^k_i$.

Propagating these $\lambda$ values up the network allows us to find all the conditional probabilities $Q^{kl}_{ij}$. These in turn can be used to calculate the means:

$$m^k_s = \sum_t \mu_{st} \sum_l Q^{kl}_{st} m^l_t \quad (9)$$

If we write $m^k_s = \alpha^k_s \pi^k_s \lambda^k_s$, then we obtain from (7)

$$\pi^k_s = \sum_{tl} \mu_{st} P^{kl}_{st} \pi^l_t \frac{\lambda^l_t}{\lambda^l_{st}}$$

with $\lambda^b_{ij}$ given by $\alpha_i \sum_k P^{kb}_{ij} \lambda^k_i / \alpha_j$.

This result is very similar to belief propagation in trees. The whole $Q$ distribution and marginals $m$ (given the $\mu$ values) can be calculated in two passes. The $\lambda$ values are propagated up the network and this gives the $Q$ distribution. Then the means $m$ can be propagated down the network. In the special case of $\mu_{ij} = 1$ for only one value of $j$, and $\mu_{ij} = 0$ otherwise (this defines a simple tree structured belief network) the above algorithm reduces to Pearl belief propagation. This can be seen by noticing that in Pearl propagation the constants of proportionality are given by $\alpha_i = \alpha_j = P(X^E)$, and by observing that with the above assumption about $\mu$,

$$\lambda^b_{ij} = \sum_k P^{kb}_{ij} \lambda^k_i \text{ and } \lambda^b_i = \prod_j \lambda^b_{ij}$$

and hence

$$\pi^a_s = \sum_{tb} P^{ab}_{ij} \pi^b_j \prod_{k \neq j} \lambda^b_{ik}$$

These are the standard Pearl propagation rules for a tree-structured belief network.

**Optimising $\mu$** The $\mu$ terms can be optimised in a similar sort of way.

$$\frac{\partial KL}{\partial \mu_{st}} = \frac{\partial V_{st}}{\partial \mu_{st}} + \sum_k \frac{\partial V^k_{d(s)}}{\partial m^k_s} \frac{\partial m^k_s}{\partial \mu_{st}} \quad (10)$$

$\frac{\partial V_{d(s)}}{\partial m^k_s}$ is the $T^k_s$ given by (6). The rest of the terms are straightforward to compute and give us

$$\mu_{st} \propto \rho_{st} \exp(-\sum_{kl} Q^{kl}_{st} m^l_t [\log Q^{kl}_{st} - \log P^{kl}_{st} + T^k_s])$$

which simplifies to

$$\mu_{st} \propto \rho_{st} \exp(\sum_l m^l_t [\log \sum_k P^{kl}_{st} \lambda^k_s]) \quad (11)$$

where the proportionality constants are obtained through normalisation.

**The complete process** Putting all this together, the procedure for optimising the parameters of the approximating distribution is as follows.

1. For each evidential node, initialise the $\lambda_i^k$ to the instantiated value of $x_i^k$.

2. Propagate the $\lambda$ values up the network using (8).

3. Calculate the conditional $Q$ distributions using (7) and propagate the means down the dynamic tree using (9).

5. Optimise the $\mu_{st}$ using (11) and the $\lambda$'s and $m$'s already calculated.

6. Repeat until suitable convergence in the KL divergence. Convergence is guaranteed because the KL divergence is reduced at each step and is bounded below by zero. The point of convergence is a local turning point of the KL divergence, because at that point all the relevant partial derivatives are zero.

## 4 Experimental comparison with loopy belief propagation

It has already been mentioned that in the single parent limit, this variational approach becomes belief propagation, and hence gives the correct exact result. This is promising, because it suggests that in situations where there is one dominant structural interpretation of the image (a peaked $Z$ distribution), then the probability distribution over the node states should be accurate.

In addition to noting the belief propagation limit case, it could well be instructive to compare the results of this variational method with the results of loopy propagation. Loopy propagation also has the same one parent limit, but gives different propagation rules.

This variational approach has many advantages over loopy propagation. First and foremost it allows us to obtain a lower bound on the log probability of the data. This can be used to learn the various parameters of the $P$ distribution. Second the variational approach provides a full joint distribution. This allows conditional and joint probabilities to be calculated. For example the probability of certain nodes for specific choices of structure can be calculated. Third, because the form and method of the approximation are explicit, it is possible to have some intuition about when the approximation will be good.

Loopy propagation has none of these features. However it is known to be good at calculating marginal probabilities, despite its approximate nature. Variational methods can be relatively poor at producing good marginals. This is because even if the joint distribution is a good approximation, this distribution corresponds to one minima of the KL divergence. Generally with this form of KL divergence, the approximation tends to be 'narrower' than the true posterior, picking out one preferred 'interpretation' of the data, but possibly ignoring others. When the marginals are calculated the other variables are only integrated out over this one interpretation, possibly skewing the resulting value for the marginal.

It is therefore a good test of this method to see how it fares in terms of calculation of the marginals. To do this a simple problem is set up. In order to make it nontrivial, we ensure that the data is chosen so that the posterior structure will not have a peaked posterior over tree structures.

A four layer network, with four nodes in each layer is defined, with each node taking one of three states. Each node in the first three layers has two possible parents in the layer above, one directly above, and one (rotationally) to the right of that. The node directly above is preferred ($\rho = 0.6$). The conditional probability tables were chosen randomly, with a strong diagonal (normalise($3I + R$) where $R$ is matrix with uniform(0,1) elements). The bottom nodes were instantiated.

Because this network is small the exact posterior can be accurately calculated through importance sampling (1000 samples were used). In addition the loopy propagation marginals and the variational marginals were calculated. A typical example of the results is given in table (1). Generally the structural variational method performs better than loopy propagation. The variational method is in fact, reasonably accurate: usually correct to within one decimal place for every probability. Marginal KL divergence calculations over a number of different runs (and structures) do confirm that the variational method does calculate better marginals, for varying conditions. The marginal KL divergences were summed over each node, and averaged over 50 different runs. The resulting average KL diverences between each approximation and the true distribution were 5.5 and 7.0 for the variational method and for loopy propagation respectively.

In conclusion then, this variational procedure does produce reasonable marginals, as well as providing the necessary joint distributions and bounds to the likelihood of any parameters of $P$.

## 5 Experimental comparison with mean field method

In order to compare the structured variational approach with the mean field method, tests were done using a 6 layer, one dimensional dynamic tree, where we represent the two possible node states by the colours black and white.

True Marginals

| 0.772 | 0.126 | 0.101 |
|---|---|---|
| 0.293 | 0.211 | 0.495 |
| 0.101 | 0.543 | 0.355 |
| 0.484 | 0.366 | 0.148 |
| 0.481 | 0.280 | 0.238 |
| 0.464 | 0.234 | 0.300 |
| 0.270 | 0.302 | 0.427 |
| 0.396 | 0.319 | 0.284 |

| Variational Marginals | | | Loopy Marginals | | |
|---|---|---|---|---|---|
| 0.742 | 0.160 | 0.097 | 0.877 | 0.063 | 0.058 |
| 0.273 | 0.268 | 0.457 | 0.399 | 0.150 | 0.450 |
| 0.067 | 0.563 | 0.369 | 0.099 | 0.482 | 0.418 |
| 0.451 | 0.390 | 0.157 | 0.488 | 0.392 | 0.119 |
| 0.450 | 0.308 | 0.241 | 0.632 | 0.208 | 0.159 |
| 0.435 | 0.279 | 0.284 | 0.566 | 0.168 | 0.265 |
| 0.252 | 0.321 | 0.425 | 0.271 | 0.260 | 0.468 |
| 0.385 | 0.328 | 0.286 | 0.384 | 0.340 | 0.274 |

Table 1: Comparison of the true marginals (for the middle two rows of the network) with those found by loopy propagation or by a DT-variational method.

150 cases of one dimensional data were independently generated from a simple Markovian process, and independent noise was added. This enables us to do a preliminary test of how well the model deals with blocky data structures. For data generated from the prior very similar results are obtained. The prior probability of the choice of a parent was given by a Gaussian decay over the distance (horizontally) between the node and the prospective parent. The positional structure of the nodes can be seen in figure 3. These Gaussians simply had standard deviations of three times the distance between the nodes on the parent layer. This ensured significant prior probability of connecting to the nearest 3 possible parents, with a small probability of connecting further afield. The conditional probability matrices had 0.9 down the diagonal and hence 0.1 on the off-diagonal.

The implementation of the mean field method was similar to that in [1] with 20 iterations of the $Q(X)$ pass for each recalculation of $Q(Z)$. For the structured variational approach, 5 passes were made through the update procedure. Any speed comparisons given here are only indicative. However because the $Q$ values can be calculated exactly in one pass, given the $\mu$ values, convergence of the structured variational approach works out to be significantly faster (convergence was assumed if the KL divergence changed by less than 0.01 on the current step).

To compare the mean field and DT-structured methods, we compare the variational free energies between the distributions found by each method and the true distribution. A lower variational free energies implies a lower KL divergence, and hence a better match to the true posterior. The results of this are in figure 2.

These results indicate that the DT-structured method gives a better approximation to the posterior than the mean field method.

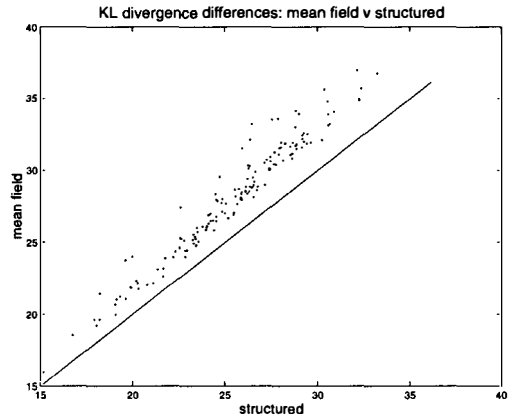

Figure 2: Comparison of the variational free energy for the DT-structured and mean field approaches

It is also instructive to look at the form of solutions which the two methods produce. First it should be noted that the mean field result is a special case of the DT-structured approximation. If the structured distribution has $Q_{ij}^{kl} = m_i^k$ for all $j, l$, it is identical the mean field distribution. When the DT-structured approach is optimised, however, we find that the $Q$ do not resemble this degenerate form. Instead they tend to be highly diagonal, and are therefore incorporating some of the conditional dependencies between the nodes.

It is interesting to see what types of structures are produced by the different methods. Figure 3 gives an example of the highest probability trees under the approximating distributions for both the mean field and the DT-structured approaches. The highest posterior tree is calculated from the distribution over tree by selecting the $Z$ structure which maximises $Q(Z)$. Note this choice is for illustration purposes only. We have (and want) posterior *distributions* over trees.

The highest probability trees for the two methods are comparable, but not always identical. For the mean field method there are problems with spontaneous symmetry breaking, which cause the higher level nodes to polarise to one or other state. This has the appearance of 'flattening out' the tree structures relating to the other state variables in the highest posterior tree (see [1] for more details). This effect can be seen in figure 2b. The nodes in the third layer from the top are all dominated by a single class. This prevents the tree structures relating to the 'black' nodes from utilising this layer. This effect is not apparent for the DT-structured approximation (figure 2a), and is a possible reason for fact that higher posterior trees are found by it.

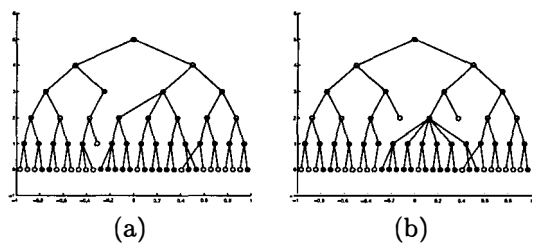

(a)           (b)

Figure 3: Comparison of the highest probability trees found by the (a) DT-structured approximation, and (b) the mean field approximation.

## 6 Conclusions

The mean field method was found to be useful for modelling dynamic trees. However it does not allow any conditional dependency structure in the distribution over the nodes. To overcome this limitation, the DT-structured variational approach has been proposed. Such an approach provides a more sophisticated model for the network node states, a model which incorporates the mean field as a special case.

The tests done in this paper suggest that the DT-structured approach succeeds in capturing the posterior distribution over the nodes better than the mean field. The evidence for this comes from both the diagonal nature of the conditional probability matrices, and the improvement in the KL divergence between the approximation and the true posterior. This variational method recovers the exact solution in the single tree limit, and hence for most real image situations, where a dominant structural interpretation would be expected, the approximation should be good.

In addition to the above, preliminary experiments suggest this approach finds accurate marginal probabilities. These marginals appear to be better than those calculated using loopy propagation, if the results of a toy demonstration are representative. The variational technique has advantages which are not available through the use of loopy propagation. These advantages are vital for the use of the dynamic tree, and include the provision of a lower bound to the log probability and the provision of joint distributions. The variational approach also provides the required *distribution* over tree structures. Finally, there are speed gains from using this method rather than the mean field. These gains come from the fact that only two order $n$ passes are needed to update the $Q$ and $m$ values; the equivalent step in the mean field involves an iterative process. The main limitation of the DT-structured approach is that the approximating distribution over the $Z$ variables still takes a factorised form.

### Acknowledgements

The work of Amos Storkey is supported through EPSRC grant GR/L78161 *Probabilistic Models for Sequences*. The author thanks Chris Williams for many helpful comments.